# Leveraging Prior Knowledge for Protein-Protein Interaction Extraction with Memory Network

Huiwei Zhou*, Zhuang Liu, Shixian Ning, Yunlong Yang, Chengkun Lang, Yingyu Lin, and Kun Ma

School of Computer Science and Technology, Dalian University of Technology, Dalian 116024, People's Republic of China

*Corresponding author: Tel: 86+411-84708140, Email: zhouhuiwei@dlut.edu.cn

## Abstract

Automatically extracting Protein-Protein Interactions (PPI) from biomedical literature provides additional support for precision medicine efforts. This paper proposes a novel memory network-based model (MNM) for PPI extraction, which leverages prior knowledge about protein-protein pairs with memory networks. The proposed MNM captures important context clues related to knowledge representations learned from knowledge bases. Both entity embeddings and relation embeddings of prior knowledge are effective in improving the PPI extraction model, leading to a new state-of-the-art performance on the BioCreative VI PPI dataset. The paper also shows that multiple computational layers over an external memory are superior to long short-term memory networks with the local memories.

Database URL: http://www.biocreative.org/tasks/biocreative-vi/track-4/

## Introduction

With the rapid growth of biomedical literature, it is becoming urgent and significant for natural language processing experts to develop entity relation extraction techniques (1-4). However, few researches have paid attention to extracting protein-protein interaction affected by mutations (PPIm) (5). The intricate networks of interactions between genes contribute to controlling cellular homeostasis, and therefore contribute to the development of diseases in specific contexts. Understanding how gene mutations and variations affect the cellular interactions provides vital support for precision medicine efforts.

For this purpose, the BioCreative VI Track 4 (5) proposes a challenging task of applying biomedical text mining methods to automatically extract interaction relations of protein pairs affected by genetic mutations, which aims to support the precision medicine initiative. There are two specific tasks in Track 4: 1) Triage task focuses on identifying scientific abstracts that describe protein-protein interaction being disrupted or significantly affected by genetic mutations; 2) Protein-Protein Interactions (PPI) extraction task focuses on extracting the affected protein pairs. This paper focuses on the PPI extraction task.

This paper presents a novel memory network-based model (MNM) for PPI extraction. The proposed model first encodes the triples (*head entity*, *relation* and *tail entity*) in Knowledge Bases (KBs) into a continuous vector space, in which a knowledge representation is learned for each entity and relation. Then, the learned knowledge representations are introduced into the memory network through attention mechanisms to capture important context clues towards a pair of entities.

Experiments on the BioCreative VI PPI dataset show that MNM could effectively leverage prior knowledge to improve PPI extraction performance. This paper also shows that multiple computational layers over an external long-term memory are crucial to state-of-the-art performance on the PPI extraction task.

## Related work

Previous researches on biomedical relation extraction mostly focus on protein-protein interactions (1-4), drug-drug interactions (6-9), and chemical-disease relations (10-14). They can be roughly divided into three categories: rule-based methods, feature-based methods and neural network-based methods.

Rule-based methods extract entity relations by adopting heuristically designed criteria (14, 15). Chen et al. (15) assume that a given protein pair contained in more than two sentences within a given document participates in a PPIm relationship. Their rule-based system achieves the highest 33.94% F1-score on BioCreative VI Track 4 PPI extraction task. Rule-based methods are simple and could achieve good performance on the specific dataset. However, it is hard to apply the extracted rules to a new dataset.

Feature-based methods (1-3, 6-8, 10, 11, 15) apply traditional machine learning techniques to learn models with one-hot represented lexical and syntactic features. Chen et al. (15) use Support Vector Machine (SVM) to learn the relation classifier with dependency features and context features. Their feature-based classifier gets the second best reported result (33.66% F1-score) in BioCreative VI Track 4 PPI extraction task. Feature-based methods need extensive feature engineering, which is time-consuming and labor intensive.

Recently, deep learning techniques have achieved great success in relation extraction tasks (4, 9, 12, 13, 16-20). Without feature engineering efforts, deep neural networks could effectively extract semantic information for relation extraction. Zeng et al. (16) first employed Convolutional Neural Network (CNN) (21) to capture the word and position information for relation extraction, and their model achieves a better performance than feature-based methods. Tran et al. (17) employ a CNN to extract local semantic features and get 30.11% F1-score on BioCreative VI PPI extraction task.

Their system achieves a relatively high precision (36.53%), but suffers from the low recall (25.61%). The reason maight be that CNN pays more attention to localized patterns and neglects global dependency information.

A number of recent efforts have been made to capture long-term information within sequences by using Recurrent Neural Network (RNN) (22) or Long Short-Term Memory (LSTM) (23) models. Zhou et al. (13) and Zheng et al. (9) both use LSTM to model long-distance relation patterns to capture the most important semantic information over a sentence.

Nonetheless, the memory in the LSTM-based models is realized through local memory cells which are locked in the network state from the past, and is inherently unstable over long timescales (24). Weston et al. (25) attempt to solve this problem by introducing a class of models called memory networks.

A memory network is a recurrent attention model with a global memory component, which allows being read and written multiple times before outputting a symbol (24). Typically, a memory network consists of a memory $m$ and four components $I$, $G$, $O$, $R$. $m$ is the input feature representation stored in the memory slot. $I$ converts the input into the internal feature representation. $G$ updates old memories with a new input. $O$ produces an output representation based on the new input and the current memory state. $R$ generates a response according to the output representation. Researches on memory networks show that the multiple computational layers over the long-term memory are crucial for good performance on the tasks of question answering (24, 25) and aspect level sentiment classification (26).

Based on the advantages of memory networks, Feng et al. (27) develop a novel attention-based memory network model for relation extraction. Their model consists of a word-level memory network which can learn the importance of each context word with regard to the entity pair, and a relation-level memory network which can capture the dependencies between relations.

All the methods mentioned above use texts as resources. Nevertheless, biomedical experts have built many large-scale KBs, which contain structured triples of protein entity pairs and their interaction relations as the form of (*head entity*, *relation*, *tail entity*), (also denoted as ($h$, $r$, $t$)), such as IntAct (https://www.ebi.ac.uk/intact/) (28), BioGrid (https://thebiogrid.org/) (29). Both two KBs have the same 45 kinds of PPI relations. Some of relations could be affected by gene mutations such as "*physical interactions*" and "*biochemical reactions*", while some other relations such as "*protein complexes*" and "*colocalizations*" are not considered in the PPI extraction task. These PPI triples provide a wealth of prior knowledge, which are crucial to PPI extraction.

How to effectively encode this prior knowledge with low-dimensional embeddings of entities and relations is an interesting topic. Recently, knowledge representation learning approach has been proposed to deal with this problem, which attempts to embed the entities and relations into a continuous vector space (30-32). TransE (30) is a typical knowledge representation learning method, which regards a relation $r$ as a translation from the head entity $h$ to the tail entity $t$ with the $\mathbf{h}+\mathbf{r} \approx \mathbf{t}$ in the embedding space, if the triple ($h,r,t$) holds. Although TransE is very simple, it could achieve state-of-the-art performance on modeling KBs (30). This paper proposes a novel memory network-based model (MNM) for PPI extraction, which employs TransE to learn embeddings of protein entities and relations from KBs. The learned knowledge representations are then introduced to two memory networks in order to capture important context clues towards a pair of entities. We show that knowledge representations significantly contribute to improving the performance, and the memory network could effectively fuse the prior knowledge and the contextual information.

## Method

Our method for the PPI extraction task can be divided into 4 steps. Firstly, the candidate instances are generated according to the pre-processing method. Then, entity-relation triples in KBs are fed into TransE model to train embeddings of entities and relations. After that, MNM is employed to capture important context clues related to knowledge representations learned from KBs for PPIm relation extraction. Finally, we apply the post-processing rules to find additional PPIm relations, and merge them with the results from MNM.

### Data

The BioCreative VI Track 4 PPI extraction task corpus contains a total of 2097 PubMed abstracts: 597 for the training set and 1500 for the test set. Proteins in the training corpus are annotated in the form of text offset and length, text span, and Entrez Gene ID, only if they participate in mutation affecting PPI relations. PPIm relations are annotated at the document level as Entrez Gene ID pair. The number of abstracts and PPIm relations in the training and test sets are listed in Table 1.

### Pre-processing

**Entity recognition and normalization** Protein entities in the training and test sets are recognized by GNormPlus toolkits (https://www.ncbi.nlm.nih.gov/CBBresearch/Lu/Demo/tmTools/download/GNormPlus/GNormPlusJava.zip) (33), and normalized to Entrez Gene ID. According to Chen et al. (15), GNormPlus have a recall of 53.4%, a precision of 40.5% and an F1-score of 46.1% for the protein name normalization task on the training set, which means not all protein mentions are annotated. For the protein mentions not annotated by GNormPlus in the training set, we can simply put them back based on the annotated protein mentions provided by the training set, and then generate the training protein pairs. But for the protein mentions not annotated by GNormPlus in the test set, we have no evidence to get them back. Table 1 lists the statistics of the test set annotated by GNormPlus ("Test-G" for short). There are totally 869 relations in the test set, of which only 483 (55.58%) are remained after protein entity recognition and 386 (44.42%) are lost since the entities in these relations cannot be recognized by GNormPlus. The low recall of entity recognition directly leads to the low recall of relation extraction.

**Table 1.** Statistics of the PPI datasets.

| Dataset | #Abstract | #PPIm |
|---|---|---|
| Training | 597 | 752 |
| Test | 1500 | 869 |
| Test-G | 1500 | 483 |

"#Abstract" and "#PPIm" mean the number of abstracts and protein-protein interaction affected by mutations in datasets, respectively.

**Candidate instances generation** Each abstract in the training set has been manually annotated with at least one relevant interacting protein pair, which is listed with the Entrez Gene ID of the two interactors. If two entities have a PPIm relation in a given document, we consider all the mentioned pairs of the two interactors in the document as positive instances.

According to the statistical results of positive instances, we use the following rules to extract candidate instances. Protein pairs not meeting these rules will be discarded.

(1) The sentence distance between a protein pair in a candidate instance should be less than 3.
(2) The token distance between a protein pair should be more than 3 and less than 50.

The same rules are applied to the Test-G set. After that, we select the words between a protein pair and three expansion words on both sides of the protein pair as the context word sequence with respect to the protein pair. To simplify the interpretation, we consider the mentions of a protein pair as two single words $w_{p_1}$ and $w_{p_2}$, where the $p_1$ and $p_2$ are the positions of the protein pair. For a given text $\{...,w_1,w_2,w_3,w_{p_1},w_{p_1+1}...,w_i,...,w_{p_2-1},w_{p_2},w_{n-2},w_{n-1},w_n,...\}$, the context word sequence we generate is expressed as $\{w_1,w_2,w_3,w_{p_1+1}...,w_i,...,w_{p_2-1},w_{n-2},w_{n-1},w_n\}$. As can be seen, we remove the mentions of the protein pair to be classified in the current instance. Then all the other protein mentions are replaced with "gene0". The numbers in the candidate instances are replaced by a specific string "NUMBER". Some special characters, such as "*", are removed.

Finally, the context word sequences of the candidate instances are acquired, which are used as the input of MNM.

### Knowledge representation learning

TransE model (30) is employed to learn knowledge representations based on the entity-relation triples in protein KBs IntAct (28) and BioGrid (29). The TransE model regards a relation $r$ as a translation from the head entity $h$ to the tail entity $t$ with the $\mathbf{h}+\mathbf{r}\approx\mathbf{t}$ in the embedding space, if the triple $(h,r,t)$ holds. TransE could learn the structure information from the triples and encode the protein entity embeddings and relation embeddings into a continuous vector space. The loss function of TransE is defined as:

$$L = \sum_{(\mathbf{h},\mathbf{r},\mathbf{t})\in S} \sum_{(\mathbf{h'},\mathbf{r},\mathbf{t'})\in S'} \max(0, \gamma - \|\mathbf{h}+\mathbf{r}-\mathbf{t}\| + \|\mathbf{h'}+\mathbf{r}-\mathbf{t'}\|) \quad (1)$$

where $\gamma$ is a margin between correct triples and incorrect triples, $S$ is the set of correct triples and $S'$ is the set of incorrect triples. KBs only contain correct triples. Conventionally, these correct triples are corrupted by replacing the head or tail entity to generate the incorrect triples $(h',r,t)$ or $(h,r,t')$. We initialize the entity embeddings with the averaged embeddings of words contained in entity mention, and the relation embeddings with a normal distribution.

Embeddings of entities and relations learned by TransE are introduced into MNM to improve the PPIm extraction performance.

### Relation extraction

To select the important context words with regard to the pair of proteins, two memory networks are adopted to pay attention to the two entity embeddings respectively. The architecture of MNM is shown in Figure 1. The two memory networks share the same parameters to learn the weight of each context word of the input sequence. Sharing the same parameters of the attention mechanisms between the two memory networks could enable the two entities to communicate with each other.

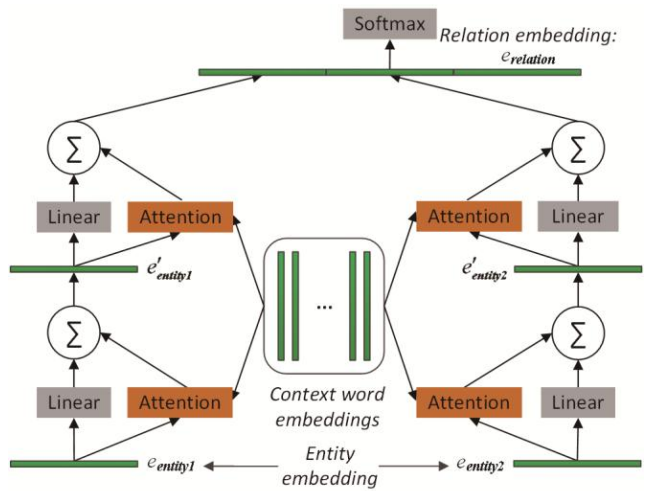

**Figure 1.** The architecture of the proposed memory network-based model. It consists of two memory networks, each of which contains two computational layers. Embeddings of entity1, entity2 and relation are learned by TransE. Note that the two memory networks share the same parameters, namely, that the same attention operation is applied to both entity1 and entity2. Finally, the two output vectors of the two memory networks and the relation embeddings are concatenated. The resulting vector is fed to the softmax layer for relation classification.

In each memory network in Figure 1, there are two computational layers, each of which contains an attention layer and a dimension-wise sum-pooling layer. The outputs of the two networks are concatenated together, and further concatenated with the relation embedding of the protein pair before given to the softmax layer for relation classification. Next, we will describe MNM in detail.

**Attention mechanism** Intuitively, not all the words in the context of a given protein pair describe the PPIm relation. For a different entity, the importance of each context word is different as well when we infer the relation of a protein pair. In this work, we employ attention mechanisms to learn the weighted score of each context word with regard to a protein pair. A higher weight indicates higher semantic relatedness with the protein pair.

In each computational layer of the two memory networks, two individual attention mechanisms are adopted to calculate the semantic relatedness of each context word with either of the two entities. The two attention mechanisms share the same parameters. Take one attention mechanism for illustration.

Given a context word sequence $s = \{w_1, w_2, ..., w_i, ..., w_n\}$ of a protein pair, the corresponding context word embeddings $\{e_1, e_2, ..., e_i, ..., e_n\}$ are regarded as the memory $m \in \mathbb{R}^{d \times n}$, where $e_i \in \mathbb{R}^d$ is a $d$-dimensional word embedding, and $n$ is the length of the context word sequence. In each attention layer, each piece of memory $m_i$ is concatenated to one entity embedding to compute its semantic relatedness with the entity. The semantic relatedness score is calculated as follows:

$$g_i = \tanh(W_a[m_i; e_{entity}] + b_a) \quad (2)$$

where $[m_i; e_{entity}]$ denotes the concatenation of memory $m_i \in \mathbb{R}^{d \times 1}$ and protein entity embedding $e_{entity} \in \mathbb{R}^{d \times 1}$, and $W_a \in \mathbb{R}^{1 \times 2d}$ and $b_a \in \mathbb{R}^{1 \times 1}$ are attention parameters. After obtaining $\{g_1, g_2, ..., g_n\}$, the attention weight of each word can be defined as follows:

$$\alpha_i = \frac{\exp(g_i)}{\sum_{j=1}^{n} \exp(g_j)} \quad (3)$$

Then the attention layer output $v_{att} \in \mathbb{R}^{d \times 1}$ is calculated as a weighted sum of each piece of memory in $m$:

$$v_{att} = \sum_{i=1}^{n} \alpha_i m_i \quad (4)$$

Though the parameters of the two attention mechanisms are the same, the weights of the same context word in the two memory networks are different, since the two concerned entities are different.

By using this attention model, semantic relatedness of each context word with a protein entity can be calculated in an adaptive way. Moreover, this attention model is differentiable, thus it can be trained easily with other components in an end-to-end fashion.

**Dimension-wise sum pooling** The output $v_{att}$ of the attention layer and the linear transformation of entity embedding are fed into a dimension-wise sum pooling layer, and the result vector is considered as a new entity embedding $e'_{entity}$ for the next computational layer:

$$e'_{entity} = W_t e_{entity} \oplus v_{att} \quad (5)$$

where $\oplus$ represents the dimension-wise sum operation, and $W_t \in \mathbb{R}^{d \times d}$ is a learned linear transformation matrix. The above sum pooling operation is applied to each of the two entities as shown in Figure 1.

Afterward, the two sum pooling vectors of the last computational layer are concatenated to form the context representations $context = [e'_{entity1}; e'_{entity2}]$. To further take advantage of the prior knowledge, relation embeddings learned from KBs are concatenated to the context representations to form output representations $output_f = [context; e_{relation}]$. Then we pass it to the softmax layer for relation classification.

**Position impact** The model mentioned above ignores the position information between context words and entities. Such position information is helpful for attention models because a context word closer to the entities should be more important than a farther one.

Following Sukhbaatar et al. (24), we control the input percentage of each piece of memory by its relative distance to the entity mention. Each percentage is calculated as follows:

$$per_i^k = (1 - p_i / n) - (k / d)(1 - 2 \times p_i / n) \quad (6)$$

where $n$ is the input sequence length, $k$ is the number of the current layer, $p_i$ is the relative distance from the current word to the entity mention, and $d$ is the dimension of word embeddings. Therefore, the actual memory vector is computed with:

$$m_i^k = e_i \odot per_i^k \quad (7)$$

where $\odot$ represents the dimension-wise product operation.

**Relation classification** The softmax layer in Figure 1 consists of a fully connected layer and a logistic regression classifier with a softmax function. It takes $output_f$ as its input and calculates the probability indicting whether the given protein pair having a PPIm relation:

$$p(y = j | T) = \text{softmax}(W_s output_f + b_s) \quad (8)$$

$$\hat{y} = \underset{y \in [0,1]}{\arg\max} \left( p(y = j \mid T) \right) \quad (9)$$

where $W_s \in \mathbb{R}^{2 \times 3d}$ is a learned transformation matrix, $b_s$ is a learned bias vector, and $T$ means all training instances.

The cross-entropy loss function is used as the training objective. For each given instance $T^{(l)}$ with its true label $y^{(l)}$, the loss function is calculated as follows:

$$loss = -\frac{1}{N} \sum_{l=1}^{N} \log p(y^{(l)} \mid T^{(l)}) \quad (10)$$

where $N$ is the number of labelled instances in the training set and the superscript $l$ indicates the $l$-th labelled instance. We adopt Adam technique (34) to update parameters with respect to the loss function.

### Post-processing

Inspired by Chen et al. (15), we use the following post-processing rules to further improve performance. If more than $N$ sentences refer to a given protein pair within a given document, the protein pair could be considered to have a PPIm relation. We set the sentence support threshold $N$ to 2 based on the statistics of the training set.

After extracting positives by the post-processing and MNM, we merge them together as final positives.

In the testing phase, there may exist multiple instances of the same protein pair in a document. It is possible that the same protein pair in a document is predicted inconsistently. If at least one instance is predicted as positive by our model of the same protein pair, we would believe this protein pair has the true PPI relation.

## Results and discussion

Experiments are conducted on the BioCreative VI Track 4 PPI extraction task corpus (35). The organizers provide 597 training PubMed abstracts, with the annotated interacting protein pairs (and the corresponding Gene Entrez IDs). The test set consists of 1,500 unannotated abstracts which are needed to recognize proteins first and then classify each protein pair into interacting or non-interacting pairs. We directly extract PPIm from biomedical documents without document triage (identifying documents that describe PPI impacted by mutations). We train our model by using the training set, and evaluate it on the test set.

Protein entities in the training and test sets are recognized by GNormPlus toolkits, and normalized to Entrez Gene ID. The evaluation of PPI extraction is reported by official evaluation toolkit (https://github.com/ncbi-nlp/BC6PM) which adopts micro-averaged (36) Precision (P), Recall (R) and F1-score (F) based on Entrez Gene ID matching (Exact Match) to measure the performance. Note that Micro performance is based on combining results from each interactor protein pair in all documents, weighting equally all pairs, regardless of the number of interactions mentioned in each document.

Word2Vec tool (https://code.google.com/p/word2vec/) (37) is used to pre-train word embeddings on the corpus (about 9,308MB) downloaded from PubMed (http://www.ncbi.nlm.nih.gov/pubmed/). The corpus consists of 27 million documents, 3.4 billion tokens, and 4.2 million distinct words. TransE model (available at: https://github.com/thunlp/Fast-TransX) is employed to learn knowledge representations. The dimension of word, entity, and relation embeddings are all 100. If a protein entity is absent in KBs, the entity embedding is initialized as an average of its constituting word embeddings. And for protein pairs not found in KBs, the corresponding relation embeddings are initialized as the zero vector. MNM is trained by using Adam technique (34) with a learning rate 0.001 and a batch size 100. All the hyperparameters are tuned to optimize model by conducting 5-fold cross-validation on the training set. The whole framework is developed by PyTorch (http://pytorch.org/).

### Prior knowledge resources

We extract PPI relation triples from KBs IntAct (28) and BioGrid (29), and they have the same 45 kinds of PPI. Since protein entities from the two KBs have different identities, we link them to standard database identifiers (Gene Entrez ID in this paper) by using UniProt (38) database. The number of triples extracted from IntAct and BioGrid is shown in Table 2. We merge these two sets of KBs by linking all protein entities to Gene Entrez IDs. Triple selection is strict without redundancy across the two resources. Finally, 1,518,592 unique triples and 84,819 protein entities are obtained for knowledge representation training as shown in Table 2.

**Table 2.** Statistics of knowledge bases.

| Knowledge bases | #Triple | #Protein entity | #Relation type |
|---|---|---|---|
| IntAct | 446,992 | 78,086 | 45 |
| BioGrid | 1,144,450 | 65,083 | 45 |
| Merged | 1,518,592 | 84,819 | 45 |

The last row named "Merged" means the combination of the triples and protein entities extracted from IntAct and BioGrid. "#Triple", "#Protein entity" and "#Relation type" mean the number of triples, protein entities and relation types, respectively.

The percentage of protein entities and relation triples covered by KBs is shown in Figure 2. From the figure we can see except the relation triple coverage on the test set, KBs cover most of the protein entities and relation triples on the training set and most of the entities on the test set.

From Figure 2, we also find that the training and test sets show different distributions. The main reason is the training set and the test set come from different sources. According to Doğan et al. (35), the training set is selected from IntAct (28), which is a specialized curation database including both interaction triples and curated articles. The interaction triples, including PPI, in the IntAct database (28) are derived from curated articles. All these PPI are extracted as prior knowledge resources in this paper. However, the test set contains only novel, not-previously curated articles (35).

Therefore, the percentage of entities and triples included in the test set is lower than that in the training set, as shown in Figure 2.

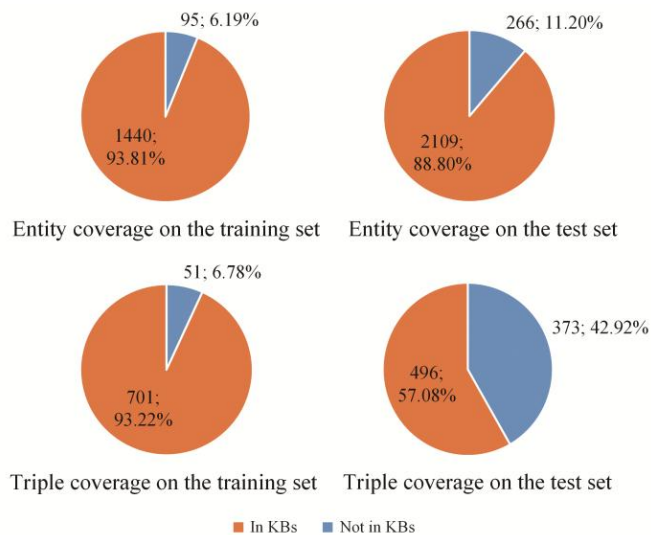

**Figure 2**. The percentage of protein entities and relation triples covered by knowledge bases. The top two panels show the entity coverage on the training and test sets respectively. The bottom two panels show the triple coverage on the training and test sets respectively.

### Effects of prior knowledge

In the experiments, we first evaluate the effects of prior knowledge. The proposed **MNM** with 4 computational layers is compared with the following baseline methods:

**AE** (**Averaged Entity Embeddings**): This method represents entity embeddings as an average of their constituting word embeddings, and directly feeds the concatenation of the two networks outputs to the softmax layer. That is to say, entity embeddings and relation embeddings learned from KBs are not used at all.

**TE** (**TransE-based Entity Embeddings**): This method employs TransE-based entity embeddings learned from KBs, and also directly feeds the concatenation of the two networks outputs to the softmax layer. That is to say, only entity embeddings learned from KBs are employed, while relation embeddings are not used.

**AE-TR** (**Averaged Entity Embeddings and TransE-based Relation Embeddings**): This method represents entity embeddings as an average of their constituting word embeddings, and feeds the concatenation of the two network outputs and the relation embeddings to the softmax layer for relation classification. That is to say, entity embeddings learned from KBs are not used, while relation embeddings are employed.

Table 3 lists the comparison results. Seen from the table, **MNM** outperforms the three baseline methods. Among the three baselines, the best one is **AE-TR**, which employs relation embeddings learned from KBs. Actually **AE-TR** is similar to **MNM** except that the entity embeddings used in **AE-TR** are not learned from KBs.

Compared with **AE**, **TE** employs entity embeddings learned from KBs, and makes the F1-score improve by 1.94%, indicating that structured knowledge information contained in TransE-based entity embeddings is more effective than the implicit semantic information expressed by word embeddings for relation classification. **AE-TR** simply adds the relation embeddings to **AE** and improves the F1-score by 2.18% compared to **AE**, which indicates that relation embeddings could provide effective clues about PPI relations to classifier. In **MNM**, both entity embeddings and relation embeddings are learned from KBs. **MNM** achieves an F1-score of 35.91%, 3.52% higher than **AE**, and outperforms the top ranked system (15) of this task.

We also show the statistical significance of the overall improvements achieved by our **MNM** over **AE**, **TE** and **AE-TR**, by using a paired *t*-test. From the results, we can see that all the improvements are statistically significant.

**Table 3.** Effects of prior knowledge.

| Prior knowledge | P (%) | R (%) | F (%) |
|---|---|---|---|
| AE | 29.86 | 35.37 | 32.38** |
| TE | 30.55 | **39.17** | 34.33** |
| AE-TR | 37.40 | 32.14 | 34.57* |
| MNM | **40.32** | 32.37 | **35.91** |

**AE**, **TE** and **AE-TR** are the variants of **MNM** that use prior knowledge or not. The marker * and ** represent *p*-value < 0.05 and *p*-value < 0.01 respectively, using paired *t*-test against **MNM**.

### Effects of architecture

To better understand our model, we study three variant architectures of **MNM**. All the variant architectures have 4 computational layers and use the same input as **MNM**.

**MNM-Single**: This is a single memory network version of **MNM**, which employs the two entity embeddings in one set of memory units rather than two separate sets of memory units. Specifically, the two entity embeddings are concatenated to each context word embedding to form the input of the attention layer.

**MNM-DA**: This architecture uses two memory networks as **MNM** does. However, different from **MNM**, the attention parameters in the two memory networks are totally different. Here, "DA" is short for "different attention".

**MNM-Max**: This method does a dimension-wise max pooling operation rather than the dimension-wise sum pooling operation at the end of each layer in **MNM**.

Table 4 shows the effects of architecture. The observations from Table 4 are listed as follows: (1) The results with two memory networks (**MNM-DA**, **MNM-Max** and **MNM**) are generally better than the results with a single memory network (**MNM-Single**); (2) The two memory networks sharing the same parameters (**MNM-Max** and **MNM**) are superior to the two memory networks using the different parameters (**MNM-DA**). From the results, we could conclude that different attention operations to the two relevant entities would introduce more noises, which are not helpful to relation classification; (3) Comparing with **MNM**, **MNM-Max** utilizes dimension-wise max pooling and causes the F1-score to drop by 0.33%. It is likely that taking the maximum value of each dimension by max pooling operation may ignore some important contextual information.

**Table 4.** Effects of architecture.

| Architectures | P (%) | R (%) | F (%) |
|---|---|---|---|
| **MNM-Single** | 41.56 | 30.07 | 34.89** |
| **MNM-DA** | 34.70 | **36.06** | 35.37* |
| **MNM-Max** | 38.83 | 32.83 | 35.58* |
| **MNM** | **40.32** | 32.37 | **35.91** |

**MNM-Single**, **MNM-DA** and **MNM-Max** are the variant versions of **MNM**. The marker * and ** represent $p$-value $< 0.05$ and $p$-value $< 0.01$ respectively, using paired $t$-test against **MNM**.

### Effects of computational layer number

In this section, we further study the effects of the number of computational layers in **MNM**. Experimental results are listed in Table 5. **MNM** with different numbers of computational layers are expressed as **MNM** ($k$), where $k$ is the number of the computational layers. When the number is less than 4, we can observe that more computational layers could generally lead to better performance. The best F1-score is achieved when the model contains 4 computational layers. When computational layer number exceeds 4, the performance becomes worse. The reason might lie in the gradient vanishing problem with the number of computational layers increasing.

**Table 5.** Effects of computational layer number.

| Computational layer number | P (%) | R (%) | F (%) |
|---|---|---|---|
| **MNM(1)** | 38.26 | 29.84 | 33.53 |
| **MNM(2)** | 39.35 | 31.91 | 35.24 |
| **MNM(3)** | 33.30 | **37.21** | 35.15 |
| **MNM(4)** | 40.32 | 32.37 | **35.91** |
| **MNM(5)** | 36.52 | 34.79 | 35.63 |
| **MNM(6)** | 34.17 | 36.18 | 35.14 |
| **MNM(7)** | **40.95** | 30.76 | 35.13 |
| **MNM(8)** | 35.93 | 35.02 | 35.47 |

The $k$ in **MNM($k$)** is the number of computational layers.

### Comparison with other methods

Table 6 compares our **MNM** with the following methods:

**CNN:** This method applies CNN with convolution, max pooling operations. In the convolution layer, 200 feature maps with window size $k = \{3, 4, 5\}$ respectively are learned. The word sequences of **CNN** are the same as **MNM**. In order to exploit the position information, this method appends two relative position embeddings to each word embedding in the sequence. And the position is defined as the relative distances from the current word to the head or tail entity.

**CNN+KB:** In addition to position embeddings, this method appends the two entity embeddings learned from KBs to each context word embedding. Finally, the concatenated representations of the max-pooling results and the relation embeddings are fed to the softmax layer.

**CNN+KB+Rule:** This method merges the protein pairs extracted by **CNN+KB** and the post-processing rules.

**Bi-LSTM:** This method applies bidirectional LSTM (23) with both word embeddings and position embeddings. In each direction, the position embedding of the current word towards one of the two entities is concatenated to each word embedding.

**Bi-LSTM+KB:** In addition to position embeddings, this method appends the two entity embeddings to each context word embedding. For the forward sequence, we concatenate one entity embedding to each word embedding; for the backward sequence, we concatenate the other entity embedding to each word embedding. And in each direction, an attention mechanism is applied to calculate the semantic relatedness of each time step hidden representation with one of the entities. Finally, the bi-directional weighted sum of the hidden representations are concatenated with the relation embeddings. The resulting vectors are fed to the softmax layer for relation classification.

**Bi-LSTM+KB+Rule:** This method merges the protein pairs extracted by **Bi-LSTM+KB** and the post-processing rules.

From Table 6, we can see that knowledge representations learned from KBs could consistently improve the performance in both CNN-based and LSTM-based methods, especially the precision of all the methods. As a complement to KBs, the post-processing can improve the recall and achieve a balance between the precision and the recall.

In addition, we find that the LSTM-based models are superior to the CNN-based models in general. This may be due to the fact that the LSTM-based models could capture long-term structure within sequences through local memory cells which are lacking in CNN-based models. After all, PPIm relations in this corpus are at document level and mainly reflected in global information.

Furthermore, the proposed MNM-based models perform significantly better than the LSTM-based models. An inherent advantage of MNM-based models lies in the external memory of memory network, which can explicitly reveal the importance of each context word in long sequential data, while LSTM can only capture the contextual information implicitly through local memory cells. Especially for the complex semantic context describing PPIm relations, explicitly extracting the important information appears to be more effective.

**Table 6.** Comparison with other methods.

| Methods | P (%) | R (%) | F (%) |
|---|---|---|---|
| **CNN** | 29.90 | 34.10 | 31.86** |
| **CNN+KB** | 36.02 | 33.37 | 34.64* |
| **CNN+KB+Rule** | 34.75 | 37.51 | 36.08* |
| **Bi-LSTM** | 27.77 | **38.13** | 32.14** |
| **Bi-LSTM+KB** | 38.75 | 31.57 | 34.79* |
| **Bi-LSTM+KB+Rule** | 36.47 | 36.18 | 36.32* |
| **MNM** | **40.32** | 32.37 | 35.91 |
| **MNM+Rule** | 37.99 | 36.98 | **37.48** |

"+KB" means using entity and relation embeddings in the corresponding model. "+Rule" means merging the protein pairs extracted by the post-processing rules and the model. The marker * and ** represent $p$-value $< 0.05$ and $p$-value $< 0.01$ respectively, using paired $t$-test against **MNM**.

### Comparison with related work

We compare our work with other related work on this task in Table 7. We also report the results evaluated on HomoloGene Match in Table 8. There may be multiple Entrez Gene IDs mapped to the same HomoloGene ID, which causes the difference between Table 7 and Table 8. Chen et al. (15) apply a rule-based approach which assumes a protein pair contained in more than two sentences within a given document participate in a PPIm relationship. This rule-based approach achieves the highest rank in the PPI extraction task. Typically, hand-crafted rules are clear and effective, but they are hard to apply to a new dataset. Chen et al. (15) also develop an SVM-based system, which uses dependency features and context features to learn relation classifier. Their feature-based system gets the second best reported result with an F1-score of 33.66%. However, feature-based methods need extensive feature engineering, which is time-consuming and labor intensive.

Apart from a traditional rule-based method and machine learning technics, Tran et al. (17) employ CNN to implicitly extract semantic features and achieve a relatively high precision. But their approach suffers from the low recall, which is caused by paying attention on localized patterns and neglecting global semantic information. Compared with these systems, our system applies memory networks to fuse contextual information with prior knowledge in KBs. Moreover, our system gets relatively balanced precision and recall after post-processing and outperforms all the systems mentioned above.

**Table 7.** Comparison with related work (Exact Match evaluation).

| Related work | P (%) | R (%) | F (%) |
|---|---|---|---|
| CNN (17) | 36.53 | 25.61 | 30.11 |
| SVM (15) | 34.49 | 32.87 | 33.66 |
| Rule-based (15) | 38.90 | 30.10 | 33.94 |
| **MNM** | **40.32** | 32.37 | 35.91 |
| **MNM+Rule** | 37.99 | **36.98** | **37.48** |

"+Rule" means merging the protein pairs extracted by the post-processing rules and the model.

**Table 8.** Comparison with related work (HomoloGene evaluation).

| Related work | P (%) | R (%) | F (%) |
|---|---|---|---|
| CNN (17) | **45.44** | 31.61 | 37.29 |
| SVM (15) | 37.61 | 35.27 | 36.40 |
| Rule-based (15) | 42.52 | 33.01 | 37.17 |
| **MNM** | 42.47 | 34.22 | 37.90 |
| **MNM+Rule** | 40.35 | **39.42** | **39.88** |

"+Rule" means merging the protein pairs extracted by the post-processing rules and the model.

**Attention visualization**

To better demonstrate the effectiveness of attention mechanism, attention weights of two example sequences are visualized in the form of heat maps in Figure 3. In Figure 3, we put the two entity mentions (located on @Entity1 and @Entity2) back to the sequence for clarity, which are removed in practice. For each sequence, the upper and the lower visible layers show the weights of the context words towards Entity1 and Entity2, respectively. In the first example, "*phosphorylation*" and "*kinase*" have the maximum weights when we pay attention to the Entity1 and Entity2, respectively. In fact, they are frequent words describing interactions according to Chen et al. (15). As for the second example, "*mutant*" and "*affinity*" have the maximum weights when we pay attention to the Entity1 and Entity2 respectively. According to Chen et al. (15), they are usually used to describe gene mutation and interactions respectively. Figure 4 lists the frequency of top 20 words being assigned the maximum weights in the sequences on the entire test set. From Figure 4, we can observe that key words such as "*phosphorylation*", "*mutant*", "*bind*", "*kinase*", "*interact*" and "*complex*" are in the list. These words are used to indicate mutations or describe interactions frequently according to Chen et al. (15). This demonstrates our **MNM** could identify the important words effectively.

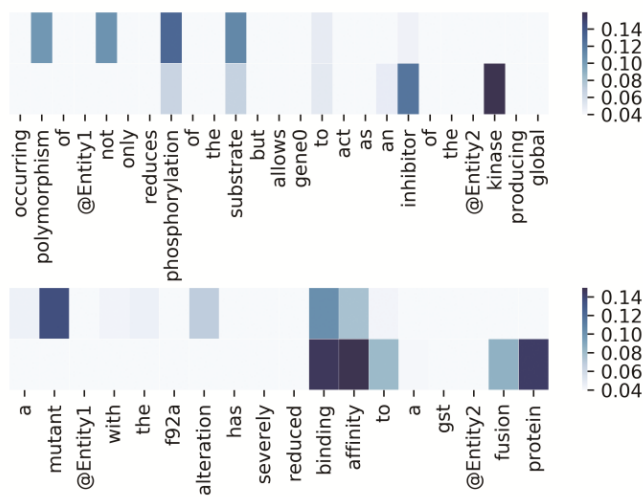

**Figure 3.** Visualization of attention weights by a heat map. Deeper color means higher weight

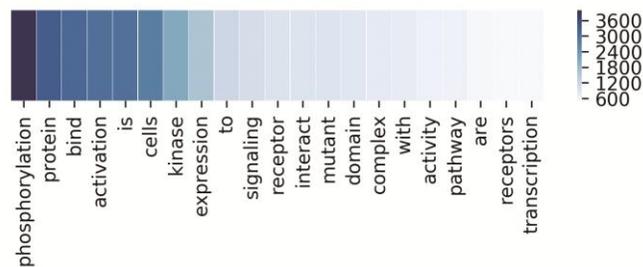

**Figure 4.** The frequency of top 20 words with the maximum weights in the corresponding sequences. Deeper color means higher frequency.

**Error analysis**

We perform an error analysis the results of **MNM+Rule** to detect the origins of false positives (FPs) and false negatives (FNs) errors, which are categorized in Figure 5 respectively.

There are two main origins of FPs (shown in the left pie chart of Figure 5):
1) Incorrect classification: 88.75% FPs are from the incorrect classification made by **MNM**, in spite of the plentiful prior knowledge and detailed contextual information.
2) Rule-based extraction error: Post-processing rules cause 11.25% of FPs.

There are three main origins of FNs (shown in the right pie chart of Figure 5):
1) False negative entity: 386 FNs with a proportion of 65.65% are caused by false negative entities without being recognized and normalized by GNormPlus toolkits, which has been mentioned in pre-processing section.
2) Incorrect classification: Due to the implicit complex semantic information of protein pairs, **MNM** misclassifies 147 positive protein pairs as negative.
3) Pre-processing error: Protein pairs distributed across more than two sentences are not extracted as candidate instances by pre-processing rules in our system, which causes 55 FNs with a proportion of 9.35%.

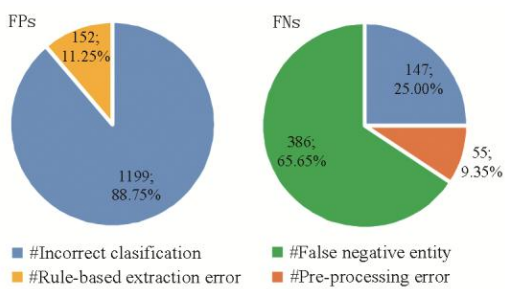

**Figure 5.** Origins of false positives (FPs) and false negatives (FNs) errors.

## Conclusions

This paper develops a novel PPIm relation extraction model with two memory networks in order to pay attention to the embeddings of protein pairs learned from KBs. The two memory networks share the same parameters and each of memory networks contains multiple computational layers. Experimental result on the BioCreative VI PPI dataset verifies that the proposed approach outperforms the existing state-of-the-art systems. This paper also shows that using multiple computational layers over an external memory is superior to LSTM with local memories.

As future work, we would like to tackle this task at a document level. In this case, how to model the whole document and select the most important information from sentence set in the document is a very challenging problem. We plan to apply a hierarchical attention network to model sentence representations with intra-sentence attention, and then model document representations with inter-sentence attention.

## Funding

This work is supported by National Natural Science Foundation of China [grant numbers 61272375, 61772109] and the Ministry of education of Humanities and Social Science research and planning Fund of China [grant numbers 17YJA740076].